\documentclass{IOS-Book-Article}

\usepackage{mathptmx}
\def\hb{\hbox to 11.5 cm{}}

\usepackage{amsmath}
\usepackage{graphicx}
\usepackage{listings}
\usepackage{xcolor}
\usepackage[normalem]{ulem}

\lstdefinestyle{prolog}{
    basicstyle=\tiny\ttfamily, 
    language=Prolog, 
    showstringspaces=false,
    commentstyle={\color{olive}},
    emphstyle={},
    breaklines=true,
    numberstyle=\tiny\color{gray},
    numbersep=10pt,
    tabsize=2,
    extendedchars=true,
    frame=trBL,
    frameround=fttt,
    frame = single,
    numbers=left,
    stepnumber=1
}

\begin{document}

\pagestyle{headings}
\def\thepage{}
\begin{frontmatter}              

\title{
Large Language Models and Explainable Law: a Hybrid Methodology
\thanks{The work has been supported by the ``CompuLaw'' project, funded by the European Research Council (ERC) under the European Union’s Horizon 2020 research and innovation programme (Grant Agreement No. 833647).}}

\markboth{}{October 2023\hb}

\author[A]{\fnms{Marco} \snm{Billi}\orcid{0000-0002-6807-073X}},
\author[A]{\fnms{Alessandro} \snm{Parenti}\orcid{0000-0002-9855-7792}}
\author[A]{\fnms{Giuseppe} \snm{Pisano}\orcid{0000-0003-0230-8212}}
and
\author[A]{\fnms{Marco} \snm{Sanchi}\orcid{https://orcid.org/0009-0005-3264-2168}%
\thanks{Corresponding Author: Marco Sanchi, marco.sanchi4@unibo.it}}

\runningauthor{M. Billi et al.}

\address[A]{University of Bologna}

\begin{abstract}
The paper advocates for LLMs to enhance the accessibility, usage and explainability of rule-based legal systems, contributing to a democratic and stakeholder-oriented view of legal technology.
A methodology is developed to explore the potential use of LLMs for translating the explanations produced by rule-based systems, from high-level programming languages to natural language, allowing all users a fast, clear, and accessible interaction with such technologies. 
The study continues by building upon these explanations to empower laypeople with the ability to execute complex juridical tasks on their own, using a \textit{Chain of Prompts} for the autonomous legal comparison of different rule-based inferences, applied to the same factual case.
\end{abstract}


\end{frontmatter}
\markboth{October 2023\hb}{October 2023\hb}

\section{Introduction}\label{s:intro}

Rule-based systems are one of the most common type of knowledge-based system used for automated legal reasoning, and they have found application in many different socio-juridical domains, such as credit approval, insurance policy determination, and public organisational structures (healthcare, welfare, pensions, etc.)\cite{lymer1995hybrid, michaelsen1984expert, masri2019survey, VANMELLE1978313}.
These systems contain a knowledge base made up of rules, often traceable to if-then statements, paired with an inference engine which applies them to factual data related to specific cases, in a transparent and explainable way.

In the domain of law, rule-based systems historically hold a well-known series of problems and limitations \cite{susskind1987expert, schauer1991playing}, including that of communicating the output resulting from their expert legal reasoning to laypeople.
Such an issue is directly related to how the knowledge base is created: to represent the domain rules in a computable way, a human expert must use specific, high-level programming languages, which ties into how the system then presents its answers to users.
The difference between natural language and these programming languages is stark, and as such it requires a way to communicate the output to stakeholders with no computer science background.
This creates a critical problem of accessibility. As the aforecited programming languages are not comprehensible by everyday users, they may not appreciate the output of the rule-based model, nor understand the justification of its legal reasoning.

The problem of communicating the complex syntax and specialised terms used in legal provisions to laypeople is not a novelty, and has been the focus of discussion ever since the sixties \cite{mellinkoff1963language, charrow1979making}, under the \textit{plain language movement}.
As a matter of fact, improving access to justice has been achieved by working on the lexical aspect of the juridical language, analyzing \textit{layperson ontologies} \cite{uijttenbroek2008retrieval, fernandez2012user}, and applying natural language processing (NLP) to simplify legal documents \cite{paquin1991loge, garimella2022text}.

A different approach towards the solution of the same problem focuses instead on 
developing understandable programming languages, such as \textit{Logical English} \cite{kowalski2020logical}.
As a Controlled Natural Language, \textit{Logical English} resembles natural language in wording, thus increasing the intelligibility of the system to the user and the programmer alike. 
However, this solution rarely takes into account the possibility for users to directly interact with the system, as these methods appear as static, and do not provide more meaningful information to different users.

In the present paper we tackle this set of issues by developing a methodology focused on employing LLMs for reprocessing the outcomes of rule-based systems in a form that is accessible to laypeople. 
Large Language Models (LLMs) are a kind of generative artificial intelligence system that leverages deep learning methodologies trained on Big Data, to achieve the processing and creation of human-like text. These models not only hold the ability to successfully generate and manipulate natural language, but also create and model programming languages, including coding script, and as such are being implemented in various fields, including the legal domain. One of the most known and used LLMs at the state of the art is GPT-4, which “\textit{exhibits human-level performance on various professional and academic benchmarks}” according to testing and research conducted by Open AI\footnote{https://openai.com/research/gpt-4}.

We argue that LLMs, by relying on the legal reasoning provided by rule-based systems, can carry out different legally relevant tasks ad present them in a form that is more accessible to end-users, compared to the one produced by the expert system. Our goal is to aid everyday users, lacking both juridical and programming skills, in appreciating the output of rule-based systems and, through the same means, in making more complex legal activities available to them.
To test these hypothesis, we provide a case study where we apply the GPT-4 model to the legal reasoning of an established rule-based system, developed using the Prolog language: CrossJustice\cite{DBLP:conf/iclp/BilliCCPSS21}. In particular, the experiments will focus on providing an accessible natural language explanation of the output of the expert system and operating a comparison between the applications of two different legal sources to the same specific case.




\section{Methodology}

In spite of their capabilities as natural language processors, LLMs have proven to struggle when they are employed for reasoning about legal norms and their applications. In these cases, such systems have shown to be plagued by hallucinations, lack of coherence and misinterpretation of the norms and the specific context \cite{savelka2023explaining, westermann2023bridging}.
For this reason, we distance ourselves from these approaches and, instead, bring about a a new way to explore the relationship between Large Language Models and rule-based systems that can enhance the access of everyday users to law.

We therefore introduce a hybrid approach where the rule-based system is employed as the legal reasoner, automatically applying the relevant norms to the specific case provided.
The rule-based system's output, presented in the form of a logic programming language, is then fed into GPT-4, for it to be reprocessed and used to carry out different operations.


\subsection{Objectives}



In developing this system, our main goal is to allow the LLM to output a result in a form that is as accessible as possible to end users, as well as providing all the necessary legal elements to enable a clear understanding of the situation at hand. 

To achieve this, we focus on one hand on giving instructions on the structure and lexicon it should be used in the answer, and on the other on ensuring that the direct relation between the specific case provided and the final outcome of the rule-based system is clearly stated in the explanation. In other words, we want the LLM to report and explain which specific conditions of the norm were applied, and exactly which facts triggered this application.

With regards to the more complex legal tasks, we aim at performing a legal analysis on different texts, in order to support both citizens and legal professionals. Our main focus is to ensure that the contrasts between legal sources were clearly highlighted and that the system could explain how these variations may be relevant for the user.
We believe all these information to be crucial for allowing the user to have a clear understanding of their legal position.


Finally, A fundamental step in our methodology is represented by repeating the experiment several times, using the same prompts on the same specific case. Given the intrinsic non-deterministic nature of LLMs, such a procedure allows us to verify whether the approach can provide correct results in a stable and consistent way.
To decrease inconsistencies, we set the model's temperature to its minimum, limiting the creativity and inventiveness of GPT-4, thus forcing it to focus on the extraction of legal inferences made by the expert system and the identification of relevant, case-based facts, to be presented in natural language. 

\subsection{Evaluation Criteria}

The results of the approach described above are evaluated according to criteria capable of validating the accessibility and legal soundness of the output. The criteria appear as follows:

\begin{itemize}
    \item \textbf{\textit{Correctness}}: accuracy in grasping key points, legal issues and essential information by the LLM. This criteria is used to exclude all output which does not match the meaning and legal argumentation of the source provided, overcoming any misinterpretations and misapplications of juridical norms, under a lenses of \textit{juridical validation}.
    \item \textbf{\textit{Form}}: coherence, readability and simplification of legal vocabulary (\textit{legalese}), to maximise the accessibility of everyday users to the output. Moreover, it verifies the correspondence between input and output in terms of structure and presentation, under the lenses of \textit{formal validation}.
    \item \textbf{\textit{Completeness}}: inclusion of all the elements requested by the prompt, with particular emphasis on those necessary to evaluate the success of operation. This criteria is used to exacerbate output that did not consider key facts about the overall process, under a lenses of \textit{substantial validation}.

\end{itemize}

\section{Case Study}


The CrossJustice platform is a rule-based system capable of automatic legal reasoning in the domain of criminal law, which provides its users with meaningful information about their rights and freedoms as suspects, or accused, of criminal conduct.
CrossJustice was identified as suitable as it holds all the main characteristics we ought to look for in order to ground our approach: it explains its inferences, it uses the high-level Prolog programming language to do so, it reasons about extremely different categories of rights, and finally it applies EU law as well as Member State law to justify its inferences.

To introduce the factual scenario provided in Listing~\ref{lst:translation_dir}, let us imagine that a person, named \textit{Mario}, is involved in criminal proceedings taking place in Poland (line 10), and does not speak the polish language (line 13). Also, he has been presented with a document charging him of his crime (line 12). According to Article 3, paragraph 2, such a document is to be considered essential (line 11). Thus, according to Article 3, paragraph 1, of the European Directive 2010/64, Mario has the right to receive a translation of this document, which is essential to ensure that he is able to exercise his right of defence and to safeguard the fairness of the proceedings. This is the main right, presented in the CrossJustice platform in lines 8-9, followed by a recap of the specific conditions needed for that right to be granted.

In the CrossJustice system we also express the relation between a primary right and the further rights that expand the meaning or the implementation of the main. 

Auxiliary rights do no directly regulate the legal sphere of the defendant, but are depending upon, or reasonably linked to a primary right. This connection can either be a temporal one, where the right exists only after the primary one has been applied, or a subjective one, which implies that the defendant has particular needs.
In this case the defendant, Mario, has the right to have the costs of the translator covered by the State under Article 4 (lines 15-23).

The property norms are used to explicitly define certain characteristics, or details, of a right. The difference between auxiliary rights and properties consists in the fact that the latter exists irrespective of the presence of the defendant, because it attains directly to the main right.
In this case, according to Article 3, paragraph 7, an oral translation of essential documents may be provided instead of a written translation on condition that such oral translation does not prejudice the fairness of the proceedings (lines 25-34).

\begin{lstlisting}[caption={Right to Translation - EU Directive}, label={lst:translation_dir}]
directive_2010_64 - art3_1

Article 3
Option: essentialDocument

Explanation:

has_right(right_to_translation, dir, art3_1, mario, essentialDocument)
    has_right(art3_1, mario, right_to_translation, essentialDocument)
        proceeding_language(mario, polish) [FACT]
        essential_document(art3_2, mario, documents)
            person_document(mario, charge) [FACT]
        not(person_understands(mario, polish))

Auxiliaries:

art4 - cost - state

Article 4
Explanation:

auxiliary_right(art4, art3_1, mario, cost, state)
    auxiliary_right(art4, mario, cost, state)

Properties:

art3_7 - form - oral

Article 3.7
Explanation:

right_property(art3_7, art3_1, mario, form, oral)
    right_property(art3_7, mario, form, oral)
        not(proceeding_event(mario, prejudice_fairness))
\end{lstlisting}

We will now explore the outcome of the same factual scenario, as applied in the Polish legal system (Listing~\ref{lst:translation_pl}). Starting from the same basic facts - Mario is involved in criminal proceedings taking place in Poland (line 10), and does not speak the polish language (line 11) - the polish legislator states that, according to Article 204, paragraph 2, of the Polish Code of Criminal Procedure, Mario has the right to translation (lines 8-9) if there is a need to translate a document drawn up in a foreign language (line 12). A document presenting a charge is one such document (line 13).

Furthermore, according to Article 618, paragraph 1, part 7, Mario has the right to have the costs of the translator covered by the State (lines 15-23).

There is no applicable article regarding the form of the translation.

\begin{lstlisting}[caption={Right to Translation - Polish Law}, label={lst:translation_pl}]
directive_2010_64_pl - article204_2

Article 204.2 code of criminal procedure
Option: documents

Explanation:

has_right(right_to_translation, pl, article204_2, mario, documents)
    has_right(article204_2, mario, right_to_translation, documents)
        proceeding_language(mario, polish) [FACT]
        not(person_understands(mario, polish))
        person_document(mario, translation_needed)
            person_document(mario, charge) [FACT]

Auxiliaries:

article618_7 - cost - state

Article 618.1.7 code of criminal procedure
Explanation:

auxiliary_right(article618_7, article204_2, mario, cost, state)
    auxiliary_right(article618_7, mario, cost, state)
\end{lstlisting}

Having established our platform of choice and case study, we are now going to focus on prompt engineering for the achievement of two different tasks. 

\section{Natural Language Translation}
\label{sec:Nat_expl}

For the first task, our goal was for GPT-4 to be capable of extracting from the Prolog trace the following pieces of information:

\begin{itemize}
    \item A simplified summary of the norms relevant to the inference;
    \item The list of rights granted to the user according to the input facts representing the case;
    \item A description of the inference process that led the system to its solution.
\end{itemize}

With these objectives in mind, we directed our efforts towards achieving the most flexible and versatile prompt, able to be applied to any of the inferences from the Crossjustice system.

There were however several problems we encountered. 
An initial challenge was to obtain a fixed structure in the output, which was often presented in different formatting (e.g. bullet lists, numbered lists, free-form text). We found that it was necessary to give a fixed structure for the LLM to follow, in order to decrease the degree of variance between answers and maximise repeatability and reliability.

Second, we required the LLM to take into consideration all legal terms and facts from the scenario, and to apply those to build the explanation, as it had a propensity to miss key facts if not well prompted.


Finally, the most challenging aspect was to overcome the tendency of the model to ascribe meaning to legal text where not explicitly mentioned nor provided. In particular, the meaning was often guessed based on the most general significance, disregarding the specificity of juridical norms and legal lexicon.

Through trial and error, we achieved the following prompt:

\begin{lstlisting}
You have been provided a Prolog inference tree using a legal norm in a specific case (Prolog Tree). 
Provide the following info according to the given structure:

Summary: simplified text of the legal norm. Use everyday langaguage with a serious register. Add any information you have on the source of law; 
What Rights do You Have: what rights do you have according to the Prolog explanation; 
Why do You Have Them: inference steps and reasoning that led to the the rights. Use all the Prolog terms in the explanation explicitly referencing the original Prolog when needed.

Use enumerations in the 'What Rights do You Have' and 'Why do You Have Them' sections if needed.
\end{lstlisting}

First, we introduced the object and the form of the input data (line 1). Second, we provided a fixed structure to follow (Summary - What Rights do You Have - Why do You Have Them). This is aimed at guaranteeing that the model follows the given formal criteria, in order to maximise repeatability and reliability.

Furthermore, we specifically requested the system to use all the Prolog terms in the explanation, with explicit references to the original (line 6). This is aimed at providing an easier way to read and evaluate the answer, both from a substantial and juridical point of view.

Finally, an example of the answers we obtained using the prompt follows:



\begin{lstlisting}
Summary: According to Directive 2010/64, Article 3.1, if you are involved in a legal proceeding and do not understand the language of the proceeding, you have the right to translation of essential documents. The cost of the translation will be covered by the state, as stated in Article 4. Additionally, under Article 3.7, you may have the right to oral translation if it does not prejudice the fairness of the proceeding.

What Rights do You Have:
1. Right to translation of essential documents.
2. State covering the cost of translation.
3. Right to oral translation, if it does not prejudice the fairness of the proceeding.

Why do You Have Them:
1. You have the right to translation of essential documents because:
   - Your proceeding language is Polish (proceeding_language(mario, polish)).
   - You have essential documents related to the charge (person_document(mario, charge)).
   - You do not understand Polish (not(person_understands(mario, polish))).
   - This right is granted under Directive 2010/64, Article 3.1 (has_right(right_to_translation, dir, art3_1, mario, essentialDocument)).
2. The state will cover the cost of translation because:
   - This auxiliary right is granted under Directive 2010/64, Article 4 (auxiliary_right(art4, art3_1, mario, cost, state)).
3. You have the right to oral translation if it does not prejudice the fairness of the proceeding because:
   - This right is granted under Directive 2010/64, Article 3.7 (right_property(art3_7, art3_1, mario, form, oral)).
   - The fairness of the proceeding is not prejudiced (not(proceeding_event(mario, prejudice_fairness))). 
\end{lstlisting}

It appears clear how GPT-4 was able to present its output in an accessible and readable way, upholding the instructions given about structure and formatting, even after much repetition (\textit{formal validation}). 
However, we cannot say that the next two criteria have been fully satisfied. GPT-4 did not include all the relevant Prolog facts used to apply juridical reasoning to the specific case (\textit{substantial validation}), as it failed to grasp and correctly represent one of the legal inferences of CrossJustice, based on the application of the the sub-rule (\textit{juridical validation}).

We noticed how GPT-4 struggles in giving the exact meaning to Prolog terms when these may be open to different interpretations. In particular, in line 11, GPT reports the fact that Mario has an essential document related to the charge as a condition of the Prolog rule. 
The predicate has been misinterpreted, and a better version would have been: ''\textit{a document containing a charge}'', or ''\textit{a document which states that the person has been charged}''. Furthermore, this has lead the LLM to mix up two facts, of which one is a condition for the application of the other. In this case, the correct solution would have been to identify that a document would be considered essential \textbf{if} that document was a charge.

The same prompt was applied to the Polish legal source:





\begin{lstlisting}
Summary: According to Article 204.2 of the Code of Criminal Procedure, if you are involved in a legal proceeding and do not understand the language of the proceeding (Polish in this case), you have the right to translation of necessary documents. Additionally, Article 618.1.7 states that the cost of this translation will be covered by the state.

What Rights do You Have:
1. Right to translation of necessary documents.
2. The cost of translation will be covered by the state.

Why do You Have Them:
1. You have the right to translation because:
   a. The proceeding language is Polish (proceeding_language(mario, polish)).
   b. You do not understand Polish (not(person_understands(mario, polish))).
   c. You have a document that needs translation (person_document(mario, translation_needed)).
   d. The document in question is a charge (person_document(mario, charge)).
   e. Therefore, you have the right to translation according to Article 204.2 (has_right(right_to_translation, pl, article204_2, mario, documents)).

2. You have the right to have the cost of translation covered by the state because:
   a. The auxiliary right in Article 618.1.7 is connected to the right in Article 204.2 (auxiliary_right(article618_7, article204_2, mario, cost, state)).
   b. Therefore, you have the right to have the cost of translation covered by the state (auxiliary_right(article618_7, mario, cost, state)).
\end{lstlisting}

Here, all 3 criteria have been fully satisfied. The performance of the LLM visibly improved, as both the sub-condition that the person Mario is presented with a document containing a charge (line 12), and that because of this Mario has a document that needs translation (line 11), have been correctly identified.
This highlights the contrast between the Natural Language translation of the same Prolog fact (\textit{''person\_document(mario, charge'')}) applied to the two corresponding legal sources, which has been interpreted differently, for no apparent reason.
These small but substantial mistakes might be a consequence of the limited context provided to the model. 
However, even when experimenting by providing the LLM the full text of the relevant legal norms, we found that it did not cause a substantial improvement in performances, nor in the language and terminology used.

Overall, we could not find a way to reliably prompt the system to correctly identify and present all sub-rules and conditions, although mistakes were significantly lowered throughout the experimenting process.

\section{Comparison of legal sources}

Building upon the results of Task 1, we followed by instructing GPT-4 to enact legal comparison between two sources. To reach a successful result, we experimented with several prompts. 
We also tested beforehand the capacity of the LLM to produce legal comparison directly on the text of the norm; however, the results were extremely poor.


We noticed that, especially for more complex tasks, employing a \textit{Chain of Thought}\cite{Wei0SBIXCLZ22, kojima2022large} 
approach decreases the probabilities of mistakes in the final answer.
\textit{Chain of Thought} prompting consists of having an LLM generate a series of intermediate reasoning steps necessary to get to the final answer. 

To implement this method, we first tried to have a single prompt describing multiple logical steps, ranging from extracting the information to the analysis of the differences.
However, we found inconsistency in the answers provided by the LLM, possibly because of the length of the step by step process and the high number of actions required.
We therefore decided to divide the prompt into two sub-prompts, each related to a specific task. 
The first one -- described in Section~\ref{sec:Nat_expl} -- requiring an explanation of both the legal sources provided, and the second one requiring their comparison based on the output of the previous step.  
%
Doing so, we created a \textit{Chain of Prompts} capable of enabling laypeople not only to better access and understand legal reasoning, but also to execute complex juridical operations such as comparison in full autonomy. 



Our goal was for the model to identify the following:

\begin{itemize}
    \item Differences and contrasts in the legal reasoning and application of the norms to the case between the sources.
    \item Diverging interpretations of the rights present in the legal sources and how those may lead to different applications of the law to the specific case.
\end{itemize}

This subsequent operation resulted in the application of the following prompt to the output obtained from the first step:


\begin{lstlisting}
You have received two legal sources. Compare them according to the following steps:
1 - COMPARE the differences in the reasonings. USE all the inference steps and describe how they influence the final outcome; 
2 - analyse potential consequences of the differences from step 1 only based on the data provided.
\end{lstlisting}

Here we followed the same principles as per the initial prompt. 
First, we introduced the object of the text given as input, although now in natural language, so no clarification regarding the form was needed.
Second, we fixed the structure to be followed when providing an answer. Again, we required the system to use all the inference steps in order to minimise the risk of misinterpretation and hallucinations. 



An example of the final results gathered using the \textit{Chain of Prompts} in discussion follows:

\begin{lstlisting}
1. Comparison of differences in reasonings:

Directive 2010/64:
- The right is granted by Article 3.1 of Directive 2010/64.
- The right is specifically for the translation of "essential" documents.
- The state covers the costs of translation as per Article 4.
- There is an additional right to oral translation if it does not prejudice the fairness of the proceeding, as per Article 3.7.

Code of Criminal Procedure:
- The right is granted by Article 204.2 of the Code of Criminal Procedure.
- The right is specifically for the translation of "necessary" documents.
- The state covers the costs of translation as per Article 618.1.7.
- There is no mention of an additional right to oral translation.

The differences in the reasonings lie in the terminology used for the documents that require translation (essential vs. necessary) and the presence of an additional right to oral translation in Directive 2010/64. The sources of the rights also differ, with one being a European directive and the other being a national code of criminal procedure.

2. Potential consequences of the differences:

- The difference in terminology (essential vs. necessary) might lead to different interpretations of which documents qualify for translation. This could potentially affect the scope of the right to translation and the number of documents that are translated for the individual.
- The presence of an additional right to oral translation in Directive 2010/64 could provide more comprehensive language assistance to the individual during the legal proceeding, ensuring better understanding and participation in the process. This right is not mentioned in the Code of Criminal Procedure, which might limit the language assistance provided to the individual.
- The difference in the sources of the rights (European directive vs. national code) could lead to potential conflicts or discrepancies in the application of these rights, depending on the jurisdiction and the specific case. This might affect the individual's access to these rights and the overall fairness of the legal proceeding.
\end{lstlisting}

With regards to the \textit{formal validation} criteria, GPT-4 was able to present its output in an accessible and readable way, upholding the instructions given about structure and formatting, even after much repetition. 

Before going into the details of the next two criteria, a clarification is necessary. 
One of the issues we faced when using a single prompt lied in the LLM's understanding of the words "\textit{documents}'' and "\textit{needed}''. We observed a tendency to ignore the context of the two words when together, and replace them with the expression "\textit{all documents}", regardless of our insistence on the use of the correct terms as extracted by the Prolog trace.
This created multiple issues from both a substantial and a juridical point of view, as it missed inference steps in its comparison, as well as misrepresented legal terms and inferences.
After changing our approach, we saw a steep increase in the quality of the outcome, as can be seen from the example above.

Currently, one of the few remaining limitations is due to the Prolog representation of the norm. In this case, GPT-4 cannot reliably infer that the Code of Criminal Procedure belongs to Poland. This is achieved in the CrossJustice platform through the use of a visual interface, while in the Prolog representation this is done through the use of the suffix \textit{\_pl}.
Furthermore, the LLM still has trouble highlighting the fact that the same document has two different interpretations according to the two applicable legal sources.

To conclude with the \textit{juridical validation} criteria, the LLM has correctly grasped the relevant terms and compared them without changing their meaning. 
However, it does not include all the relevant inference steps used to apply juridical reasoning to the specific case (\textit{substantial validation}).










\section{Conclusions}\label{s:conc}

This paper explores the opportunities and limitations regarding the use of LLMs for the autonomous generation of accessible  natural language explanations, within the context of rule-based systems in the legal domain. Moreover, it tackles the possibility of building upon these explanations to empower stakeholders with the ability of enacting autonomous legal tasks, such as comparing the application of different norms to the same specific case.

To reach our first objective, we provided a methodology for the engineering of flexible prompts, able to process juridical inferences in a stable, repeatable and simplified way. 
We followed by successfully applying our hybrid approach to the CrossJustice platform - a system based in Prolog language for the domain of criminal law - showing our method to be effective in making the rule-based reasoning accessible, while preserving its substantial, juridical and formal validity.

After establishing such sound foundation, we moved to our second objective by creating a chain of prompts able to process different rule-based outputs and their explanations. Our goal was to produce legal comparison by identifying the relevant juridical and factual differences present amongst inferences relating to the same specific case.
The methodology proved to be once again successful, showing the potential of a hybrid approach based on the expert reasoning of rule-based systems, paired with the versatility of LLMs, which opens the door to various legal operations, as shown in the case study.

On this note, future works would expand our methodology by further developing the \textit{Chain of Prompts} used in our trials, exploring the potential \textit{modularity} of such an approach. 
By this, we mean the creation of a versatile and flexible initial prompt for the natural language translation and explanation of rule-based inferences, followed by the engineering of multiple different and subsequent prompts, each dedicated to a different legal operation. Those would be applied accordingly to the output of the first, creating chains going beyond normative comparison, enabling more complex and differentiated operations.

Finally, our methodology could easily embrace the multilingual nature of European Law given the capabilities of state-of-the-art LLMs, thus contributing to overcome language barriers in the fruition of legal technology, as well as to bolster access to European and Member State law.





\bibliographystyle{vancouver}
\bibliography{main}

\end{document}